\documentclass[letterpaper, 10 pt, journal, twoside]{IEEEtran}
\usepackage{times,multirow,caption,float}
\usepackage{enumitem}
\usepackage{wrapfig}
\usepackage{graphicx}
\usepackage{epsfig,xspace,layout}
\usepackage{subfigure}
\usepackage{color}
\usepackage{amsfonts}
\usepackage{times}
\usepackage{amssymb}
\usepackage{hyperref}
\usepackage{amsmath,bm}
\usepackage{rotating}
\usepackage{amsthm}
\usepackage{graphicx}
\usepackage{epsfig}
\usepackage{mathrsfs}
\usepackage{caption}
\usepackage{subfigure}
\usepackage{mathtools}
\usepackage{makeidx}
\usepackage{multirow} 
\usepackage{dblfloatfix}
\usepackage{threeparttable}
\usepackage{dsfont}
\usepackage{cite}
\usepackage{xcolor}
\usepackage[font={small}]{caption}
\usepackage{algorithm}
\usepackage{algorithmic}
\usepackage{tikz}
\usetikzlibrary{shapes,arrows,positioning,calc}
\usepackage{siunitx}
\usepackage{multicol,lipsum}

\usepackage{url}
\usepackage{booktabs}
\usepackage{lineno}
\usepackage{hhline}
\usepackage[Symbol]{upgreek}

\theoremstyle{definition}

\theoremstyle{remark}
\newtheorem{remark}{Remark}

\DeclareMathAlphabet{\mathpzc}{OT1}{pzc}{m}{it}

\DeclareFontFamily{U}{jkpmia}{}
\DeclareFontShape{U}{jkpmia}{m}{it}{<->s*jkpmia}{}
\DeclareFontShape{U}{jkpmia}{bx}{it}{<->s*jkpbmia}{}
\DeclareMathAlphabet{\mathfrak}{U}{jkpmia}{m}{it}

\DeclareMathOperator{\blkdiag}{Blkdiag}
\DeclareMathOperator{\dt}{dt}

\renewcommand{\u}{\mathbf{u}}

\newcommand{\q}{ \bm q}
\renewcommand{\H}{ \bm H}

\newcommand{\y}{\mathbf{y}}%

\newcommand{\0}{\mathbf{0}}
\newcommand{\p}{{\bf p}}%
\newcommand{\e}{{\bf e}}%
\newcommand{\x}{\mathbf{x}}%

\newcommand{\A}{\bm A}
\newcommand{\B}{\bm B}
\renewcommand{\d}{\bm d}

\newcommand{\D}{\mathcal{D}}
\newcommand{\C}{\bm C}

\newcommand{\U}{\mathcal{U}}

\newcommand{\I}{\mathcal{I}}

\newcommand{\G}{\bm G}

\newcommand{\J}{\mathcal{J}}

\newcommand{\Q}{\mathcal{Q}}

\renewcommand{\u}{\bm u}

\newcommand{\ar}[1]{\left[\begin{array}#1\end{array}\right]}
\newcommand{\al}[1]{\begin{align}#1\end{align}}
\newcommand{\eq}[1]{\begin{equation}#1\end{equation}}
\newcommand{\ald}[1]{\begin{aligned}#1\end{aligned}}

\newcommand{\eqn}[1]{\begin{equation*}#1\end{equation*}}

\newcommand{\subeq}[1]{\begin{subequations}#1\end{subequations}}

\title{Underactuated Motion Planning and Control for Jumping with Wheeled-Bipedal Robots}

\author{Hua Chen$^{1}$, Bingheng Wang$^{2}$, Zejun Hong$^{1}$, Cong Shen$^{1}$ Patrick M. Wensing$^{3} $ and Wei Zhang$^{1}$
\thanks{Manuscript received: August, 12, 2020; Revised November 12, 2020; Accepted December, 7, 2020.}
\thanks{This paper was recommended for publication by Editor Dezhen Song upon evaluation of the Associate Editor and Reviewers' comments. This work was supported by National Natural Science Foundation of China (Grant No. 62073159, and Grant No. 62003155), and in part by the Shenzhen Science and Technology Program (Grant No. JCYJ20200109141601708).\emph{(Corresponding author: Wei Zhang.)(Hua Chen, Bingheng Wang and Zejun Hong contributed equally to this work.)}}
\thanks{$^{1}$Hua Chen, Zejun Hong, Cong Shen and Wei Zhang are with the Department of Mechanical and Energy Engineering, Southern University of Science and Technology, Shenzhen, 518055, China. {\tt \small {chenh6@sustech.edu.cn, \{hongzj, 11510411\}@mail.sustech.edu.cn, zhangw3@sustech.edu.cn}}}
\thanks{$^{2}$Bingheng Wang is with the Department of Electrical and Computer Engineering, National University of Singapore, Singapore 117583, Republic of Singapore. {\tt \small wangbingheng@u.nus.edu}}
\thanks{$^{3}$Patrick M. Wensing is with the Department of Aerospace and Mechanical Engineering, University of Notre Dame, Notre Dame, IN 46556, USA. {\tt \small pwensing@nd.edu}}
\thanks{Digital Object Identifier (DOI): see top of this page.}}
\renewcommand{\q}{{\bm q}}
\renewcommand{\H}{{\bm H}}
\renewcommand{\C}{{\bm C}}
\renewcommand{\p}{{\bm p}}
\renewcommand{\D}{{\bm D}}

\renewcommand{\J}{{\bm J}}
\newcommand{\llambda}{{\boldsymbol \lambda}}
\newcommand{\xxi}{\boldsymbol{\xi}}
\newcommand{\ttau}{\boldsymbol{\tau}}
\newcommand{\Phii}{\boldsymbol{\Phi}}

\begin{document}

\maketitle

\begin{abstract}
This paper studies jumping for wheeled-bipedal robots, a motion that takes full advantage of the benefits from the hybrid wheeled and legged design features. A comprehensive hierarchical scheme for motion planning and control of jumping with wheeled-bipedal robots is developed. Underactuation of the wheeled-bipedal dynamics is the main difficulty to be addressed, especially in the planning problem. To tackle this issue, a novel wheeled-spring-loaded inverted pendulum (W-SLIP) model is proposed to characterize the essential dynamics of wheeled-bipedal robots during jumping. Relying on a differential-flatness-like property of the W-SLIP model, a tractable quadratic programming based solution is devised for planning jumping motions for wheeled-bipedal robots. Combined with a kinematic planning scheme accounting for the flight phase motion, a complete planning scheme for the W-SLIP model is developed. To enable accurate tracking of the planned trajectories, a linear quadratic regulator based wheel controller and a task-space whole-body controller for the other joints are blended through disturbance observers. The overall planning and control scheme is validated using V-REP simulations of a prototype wheeled-bipedal robot.

\end{abstract}

\begin{IEEEkeywords}
Optimization and Optimal Control; Whole-Body Motion Planning and Control; Underactuated Robots
\end{IEEEkeywords}

\section{Introduction}\label{sec:intro}

\IEEEPARstart{B}{y} integrating the advantage of high energy efficiency from wheeled robots and the capability of dealing with sophisticated terrains from bipedal robots, wheeled-bipedal robots are able to accomplish various agile and versatile locomotion tasks. During recent years, multiple hardware platforms for wheeled-bipedal robots have been developed~\cite{Li2018,Klemm2019,Li2019,Stilman2009,Kawaharazuka2018}. However, in contrast to the rapid growth of hardware advancements, planning and control strategies for complicated motions with wheeled-bipedal robots have yet received adequate investigation. 

Generally speaking, hierarchical schemes have played an important role in controlling complex robotic systems to achieve sophisticated behaviors~\cite{Full1999,Kuindersma2016,Wensing2017}. For wheeled-bipedal robots, reliable balancing and velocity tracking controllers serve as a foundation of advanced planning and control schemes for more complicated motions. Classical proportional-integral-derivative (PID) and linear quadratic regulator (LQR) controllers have been developed for achieving this fundamental task~\cite{Liu2019,Klemm2019}. More recently, borrowing ideas from the legged locomotion literature, several whole-body control strategies for wheeled-bipedal robots have been developed to partially account for the coupling between the dynamics of the upper-body and wheels~\cite{xin2020online,Zambella2019,Zafar2019,Klemm2020,Zhou2019}. However, many of these existing studies focus on balance control problems during stance phase and have not adequately considered the associated planning problems, limiting potential motions achievable with wheeled-bipedal robots.

\begin{figure*}[htbp!] 
\centering
\subfigure[\footnotesize BD `Handle'  (2017) \newline Standing height: $\approx$198 cm]{%
\includegraphics[height=0.25\linewidth]{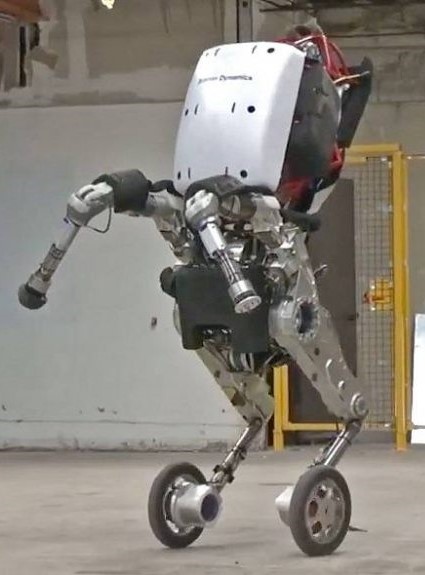}
\label{fig:handle}}
\subfigure[\footnotesize ETH `Ascento' (2018) \newline Standing height: $\approx$60 cm]{%
\includegraphics[height=0.25\linewidth]{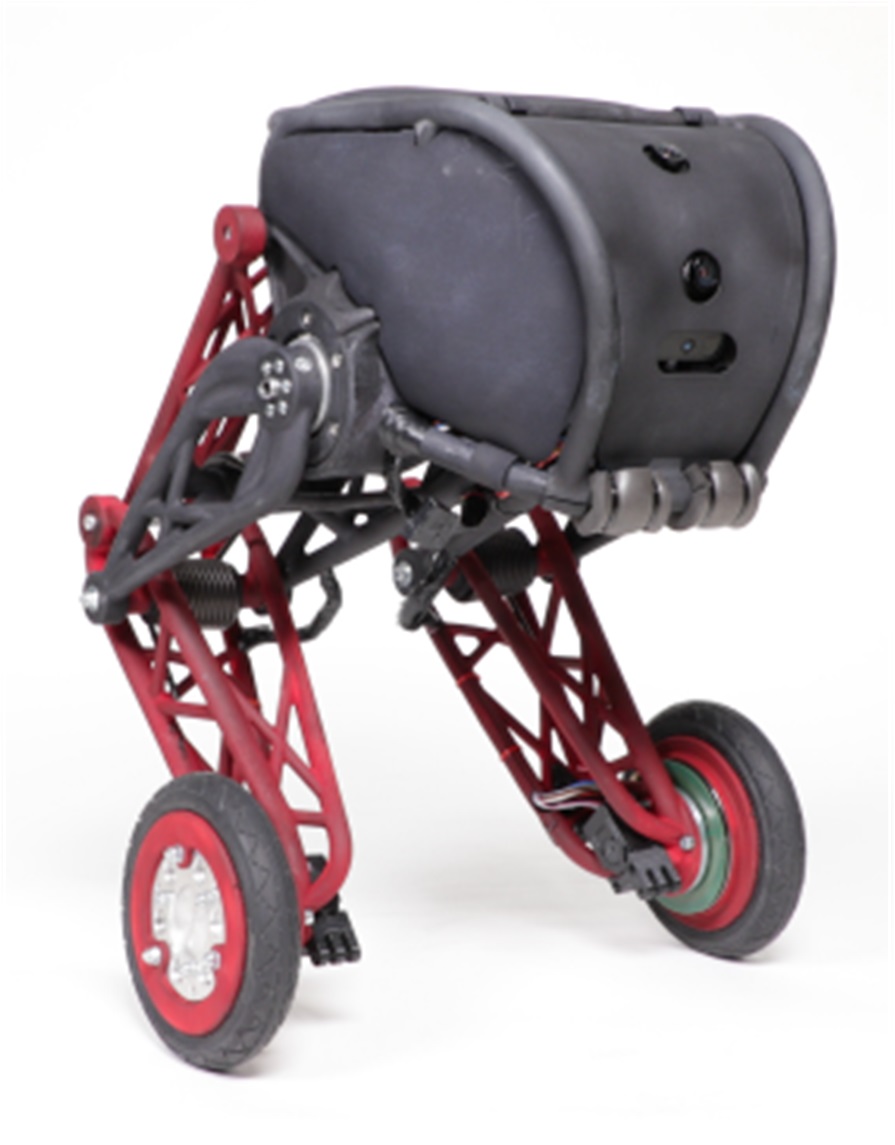}
\label{fig:ascento}}
\subfigure[\footnotesize HIT `SR600' (2019) \newline standing height: $\approx$80 cm]{%
\includegraphics[height=0.25\linewidth]{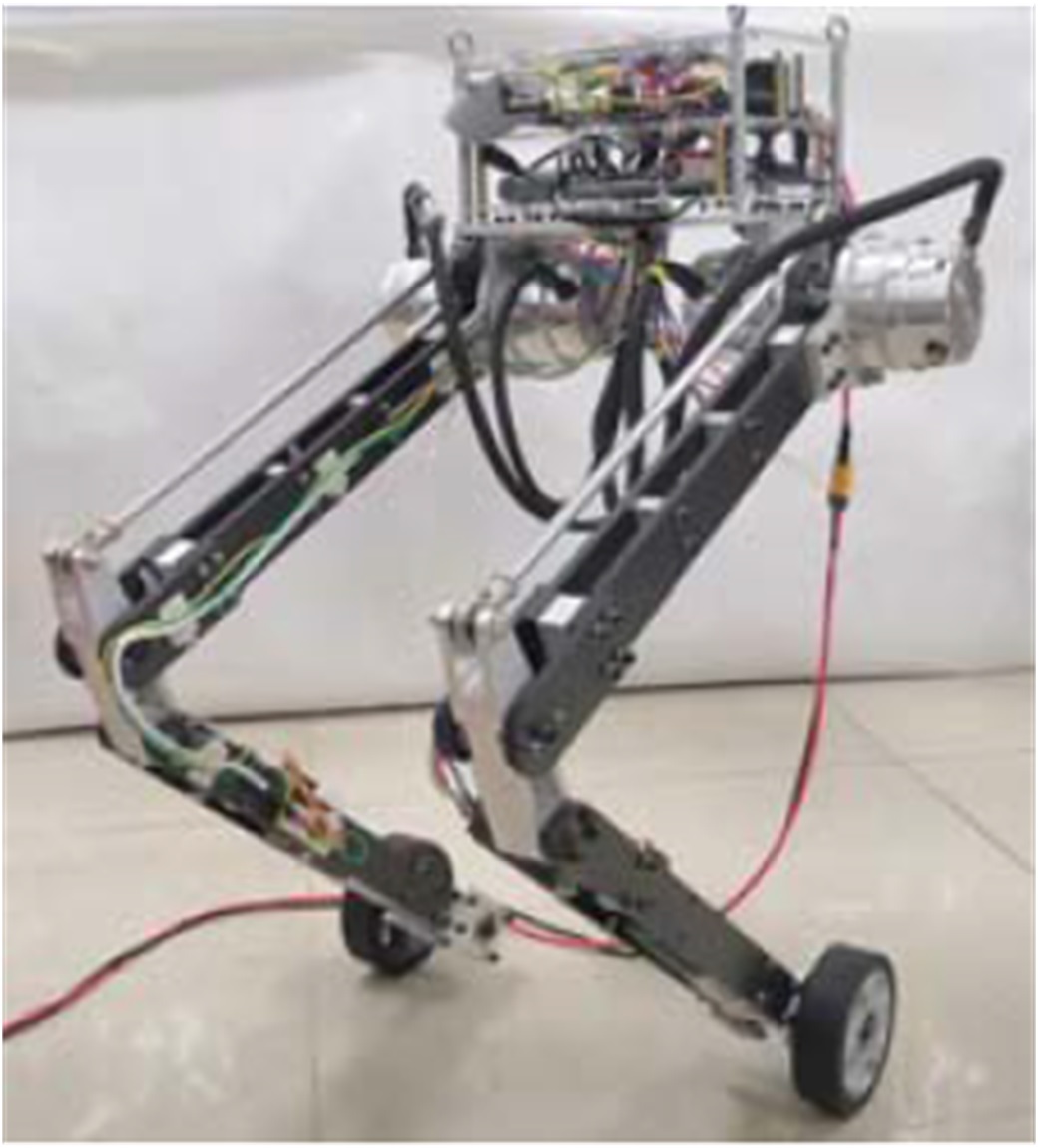}
\label{fig:sr600}}
\subfigure[\footnotesize SUSTech `NeZha' (2020)  \newline Standing height: $\approx$89 cm]{%
\includegraphics[height=0.25\linewidth]{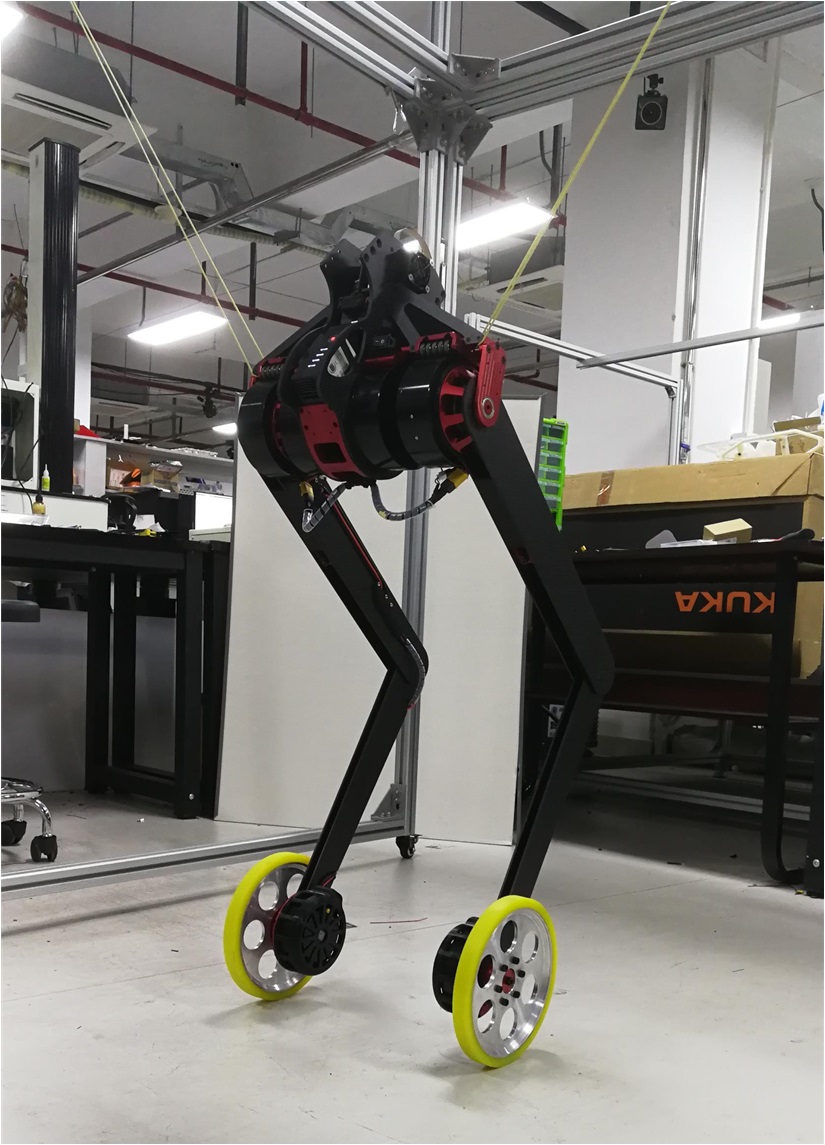}
\label{fig:nezha}}
\caption{\footnotesize Different wheeled-bipedal robots platforms, shown in chronological order. }
\label{fig:wheeled-bipedal-robots}
\vspace{-15px}
\end{figure*}

Motion planning plays a crucial role in finding feasible trajectories for robots to perform highly agile and dynamic motions~\cite{LaValle2006}. For wheeled-bipedal robots, in addition to the common challenges such as high dimensionality and nonlinearity of the underlying system, the underactuated nature of the robot's dynamics further complicates the motion planning problem. For planning of general underactuated mechanical systems, Shammas et al. proposed a geometric approach for designing trajectories in the actuated configuration space to adjust the trajectories in unactuated degrees of freedom~\cite{Shammas2007}. This approach was later extended to problems with non-holonomic constraints~\cite{Shammas2007a}. A partial feedback linearization approach has been widely adopted for control of underactuated systems~\cite{Spong1994,Spong1998,Shiriaev2014}. When combined with sampling-based planners, this approach has been shown effective for several high-dimensional underactuated systems~\cite{Shkolnik2008}. Despite these underactuated motion planning schemes, there is little related literature on planning for underactuated wheeled robots. 


In this paper, we focus on jumping with wheeled-bipedal robots, which is a representative motion that fully exploits the hybrid design morphology, allowing robots to overcome challenging terrains such as obstacles and gaps whose sizes are larger than the wheel radius while maintaining high energy efficiency. Jumping with wheeled-bipedal robots is a fundamentally challenging problem mainly due to the transitions between different dynamics during the motion and the underactuated nature of the unstable stance phase dynamics. Throughout the history of bipedal and humanoid robots, template-based planning and control strategies for jumping have been studied and validated via whole-body simulations and even real-world experiments~\cite{Poulakakis2009,Wensing2014,Xiong2018}. Recently, Dinev et al.~\cite{Dinev2020} considers the jumping problem for a simple wheeled robot with a prismatic actuator, in which a nonlinear model predictive control problem is formulated and solved to accomplish the desired motion. Despite this pioneering work, comprehensive planning and control schemes for jumping with wheeled-bipedal robots have not been reported.

The contributions of this paper are summarized as follows. First, a novel wheeled-spring-loaded inverted pendulum (W-SLIP) model is proposed as the template model for the planning problem. Such a model extends the classical spring-loaded inverted pendulum (SLIP) model to include a wheel with non-trivial mass. Inspired by the fact that the SLIP model is the most commonly adopted template model for planning jumping motion with bipedal robots, and the fact that the key difference between wheeled-bipedal robots and bipedal robots lies in the newly added wheels, the proposed W-SLIP model captures the essential dynamics of the wheeled-bipedal robots during jumping, while preserving its underactuated feature. Second, we develop a tractable quadratic programming based planning algorithm to tackle the challenging stance phase planning problem. We show that the W-SLIP model inherits a desirable differential-flatness like property, which allows for simplification of the challenging motion planning problem due to high nonlinearity and underactuation of the underlying W-SLIP dynamics. Exploiting this property, a quadratic programming based algorithm is devised, which can be efficiently solved in real time that enables online implementation in a receding-horizon fashion. Third, we propose a disturbance-observer (DOB) based composite controller for the tracking problem during stance phase. By leveraging such a DOB based composite controller for wheeled-bipedal robots, control of the fully actuated upper-body can be decoupled from that of the underactuated wheels, with the complex coupling terms being handled by the DOB. As compared with existing whole-body controllers that require careful analysis to account for the underactuated nature of the wheeled-bipedal dynamics, this DOB based strategy offers a simple yet effective scheme for controlling the underactuated robot.  Finally, all the proposed planning and tracking control schemes are integrated into a comprehensive hierarchical framework, which coordinates different planners and controllers at different timescales, enabling online implementation of the overall framework. Effectiveness of the proposed solution framework is demonstrated via simulation validations with a prototype wheeled-bipedal robot.


\section{Problem Description and Hierarchical Planning and Control Architecture}\label{sec:problem}

\begin{figure}[bp!]
\centering
\subfigure[\footnotesize Whole-body model]{%
\includegraphics[width=0.33\linewidth]{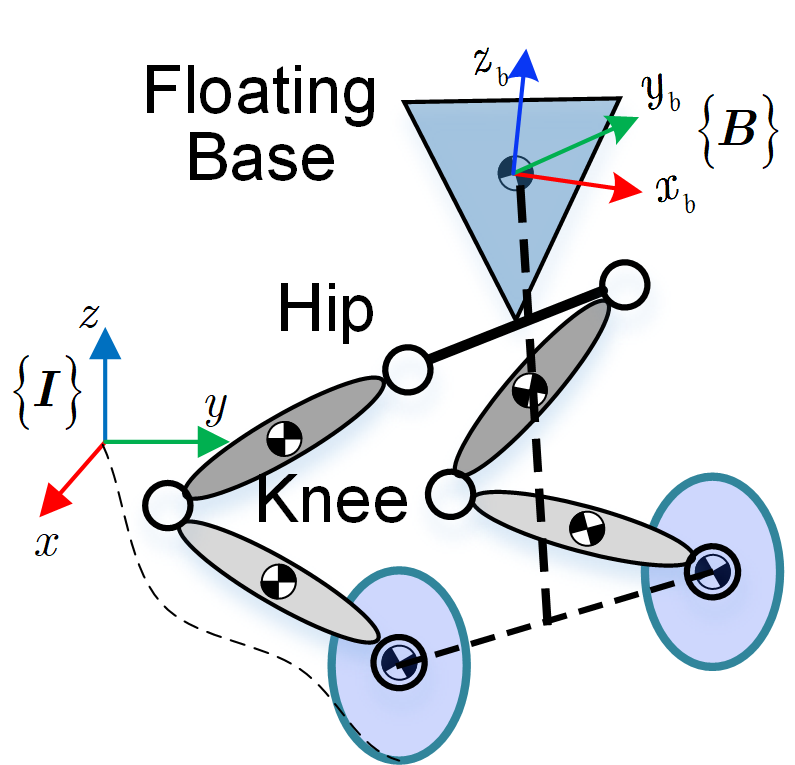} 
\label{fig:wholebody_model}}
\subfigure[\footnotesize W-SLIP in stance]{%
\includegraphics[width=0.26\linewidth]{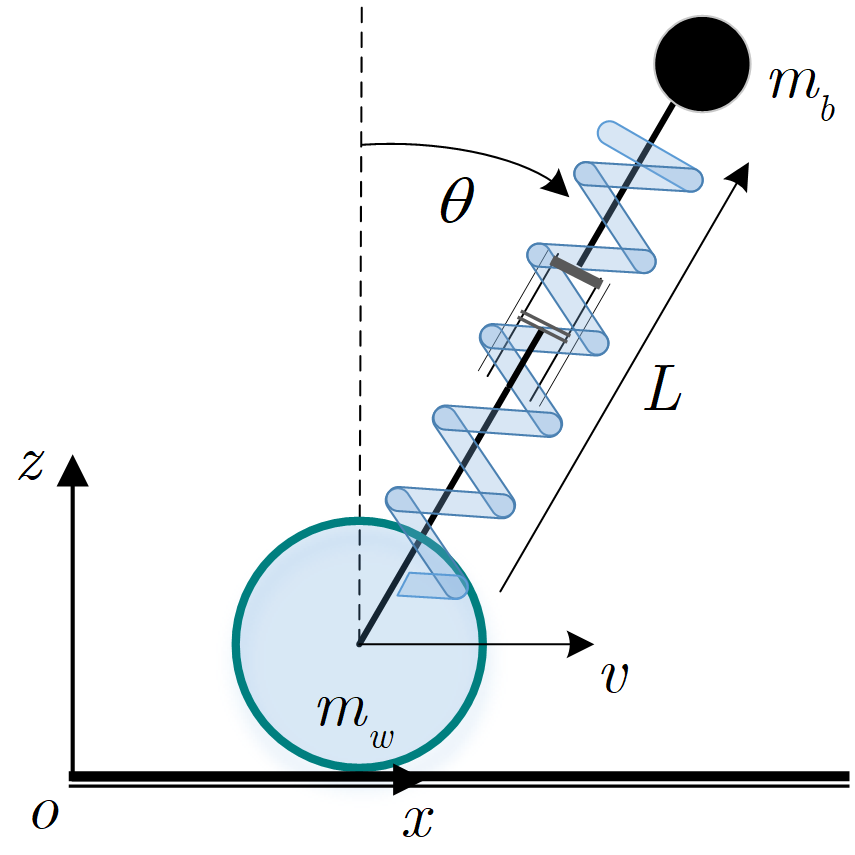} 
\label{fig:wslip_s}}
\subfigure[\footnotesize W-SLIP in flight]{%
\includegraphics[width=0.25\linewidth]{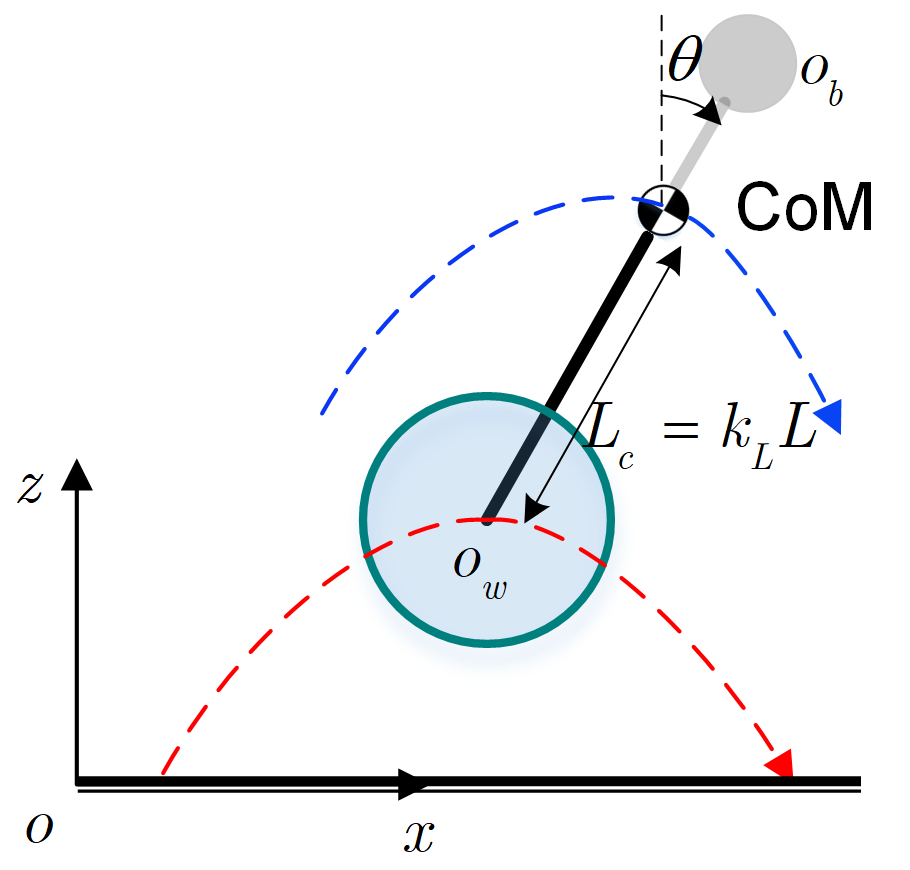}
\label{fig:wslip_f}}
\caption{\footnotesize Whole-body model and proposed W-SLIP model}
\label{fig:W-SLIP}
\end{figure}

This paper focuses on planning and control of jumping for a general class of wheeled-bipedal robots as shown in Fig.~\ref{fig:wheeled-bipedal-robots}, whose underlying dynamics can be modeled as a floating-base multi-link rigid-body system, as depicted in Fig.~\ref{fig:wholebody_model}. Denoting by $\q = (\q_\text{fb},\q_\text{J})$ the configuration of the robot containing both the floating base configuration $\q_\text{fb}$ and the joint configurations $\q_\text{J}$, and by $\boldsymbol{\tau}$ the vector of all joint torques, the generic model for wheeled-bipedal robots is given as
\subeq{\label{eq:wholebody}\al{
\H\!\left( \q \right)\ddot \q + \C\!\left( {\q,\dot \q} \right) &= {{\bm S}^T}{\boldsymbol \tau}  + \Phii^T\!\left( \q \right)\llambda \label{eq:wholebody1} \\ \Phii\!\left( \q \right)\dot \q &= 0\label{eq:wholebody2}}}
where $\H$ is the generalized inertia matrix, $\C(\q,\dot{\q})$ is the term addressing effects of Coriolis force, centrifugal force and gravitational force, ${\bm S}$ is the control selection matrix, $\llambda$ is the Lagrange multiplier of constraint forces, and $\Phii$ is used to enforce constraints due to wheel contact. During flight, the contact breaks and hence the constraint~\eqref{eq:wholebody2} and $\Phii^T\!\left( \q \right)\llambda$ in~\eqref{eq:wholebody1} representing the constraint force both vanish.

Jumping with wheeled-bipedal robots requires finding appropriate joint torque control inputs such that a flight phase is created between two stance phases. Underactuation of the unstable stance phase dynamics is one of the main difficulties to be addressed, in addition to the traditional challenges due to the nonlinearity of the stance phase dynamics and the hybrid nature of discrete transitions between stance and flight. 
\begin{figure*}[tp!]
	\centering
	{\includegraphics[width=0.8\linewidth]{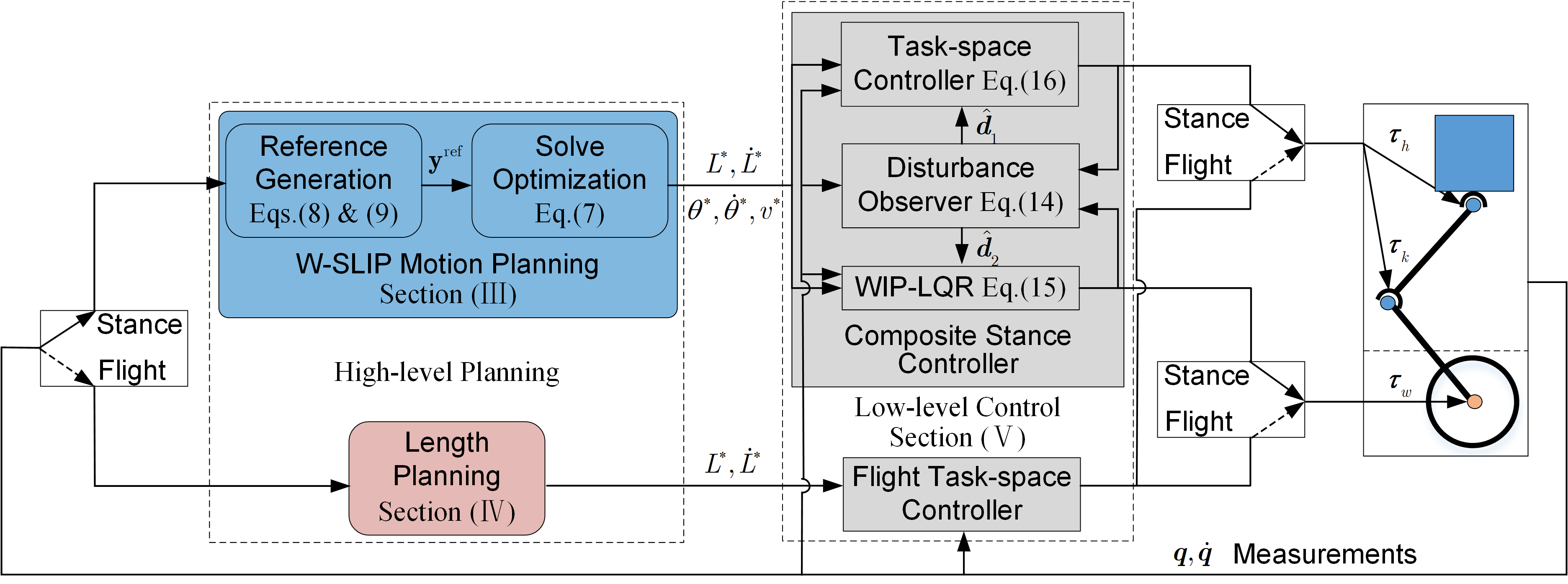}}
	\caption{\footnotesize Overall planning and control architecture proposed in this paper. Blocks shaded in blue run at $25$ Hz, blocks shaded in gray run at $500$Hz, block shaded in red runs only once at the take-off moment. }
	\label{fig:control framework}
	\vspace{-15px}
\end{figure*}
Before presenting technical details, we first provide a brief overview of the proposed hierarchical planning and control framework (Fig.~\ref{fig:control framework}), which consists of four main parts including stance phase planning, flight phase planning, stance phase control, and flight phase control. In both planning problems, a novel wheeled-spring-loaded inverted pendulum (W-SLIP) model is adopted. Based on such a model, quadratic programming based planning methods are developed, leading to feasible motions during both phases. For stance phase tracking control, a composite controller consisting of a time-varying linear quadratic regulator determining the wheel torques and a task-space controller specifying torques at the upper-body joints is devised. The task-space control method is also applied for the flight phase tracking problem, which completes the overall framework. Detailed discussions of each part of the framework will be given in later sections. 

\section{Planning for Stance Phase}\label{sec:stanceplanning}
To allow for effective motion planning for wheeled-bipedal robots, this section first proposes a simplified model describing the essential underactuated dynamics. Then, a tractable algorithm solving the motion planning problem of the underactuated template dynamics is constructed. 

\subsection{Wheeled Spring-Loaded Inverted Pendulum Model}
Motivated by the classical spring-loaded inverted pendulum (SLIP) model for abstracting bipedal robot dynamics during jumping, a wheeled SLIP (W-SLIP) model is proposed. The model (Fig.~\ref{fig:W-SLIP}) attaches a wheel with non-zero mass to the SLIP, abstracting the essential jumping dynamics of wheeled-bipedal robots. The rest length of the spring is denoted as $L^0$, with the stiffness of the spring $\bar{K}_s$, mass of the floating base $m_b$, mass of the wheel $m_w$, and radius of the wheel $r$. The rotational inertia of the wheel is considered to be negligible.

To enable control of the W-SLIP model, a linear actuator capable of modulating the spring force is considered in series with the spring and a rotary actuator able to apply torque about the wheel axis is considered on the wheel. Denote by $\q = (x,L,\theta)$ the configuration variables of the W-SLIP model representing the horizontal position, leg length and leg angle, and by $\u = (\tau_w, \Delta L)$ the torque about the wheel axis and linear actuator displacement, respectively. Equations of motion of the W-SLIP model during stance can be obtained via classical Lagrangian methods, as given below 
\eq{\label{eq:eom_wslip} \H (\q)\ddot{\q} + \C(\q,\dot{\q}) = \B \u ,} where
\eqn{\H(\q) = \ar{{ccc} m_b+m_w & m_b\sin \theta & m_b L \cos \theta \\   m_b \sin \theta  & m_b & 0 \\ m_b L\cos \theta  & 0 &   m_b L^2}} 
is the generalized inertial matrix, 
\eqn{\C(\q,\dot{\q}) = \left[ {\begin{array}{*{20}{c}}
{2{m_b}\dot L\dot \theta \cos \theta  - {m_b}L{{\dot \theta }^2}\sin \theta }\\
{ - {m_b}L{{\dot \theta }^2} + {m_b}g\cos \theta  - {\bar{K}_s}\left( {{L^0} - L} \right)}\\
{2{m_b}L\dot L \dot \theta  - {m_b}gL\sin \theta }
\end{array}} \right]} 
is the term addressing the Coriolis force, Centripetal force, and gravitational force, and 
\eqn{\B = \left[ {\begin{array}{*{20}{c}}
{{r^{ - 1}}}&0\\
0&\bar{K}_s\\
{ - 1}&0
\end{array}} \right]}
is the matrix specifying how inputs affect the dynamics.

This proposed W-SLIP model greatly simplifies the whole-body dynamics and keeps the essential underactuated feature of wheeled-bipedal robots. In specific, the wheel is capable of continuously repositioning during stance to adjust the direction of contact forces, but at the expense of short-term postural destabilization due to necessary rolling torque applied at the wheel axis. To account for the underactuated feature, a quadratic programming based planning scheme for the W-SLIP model is developed in the sequel.

\subsection{Planning for Underactuated W-SLIP Model}
Planning for the W-SLIP model during stance is cast as the following optimal control problem.
\subeq{\label{eq:oc_planning_stance}\al{ \min\limits_{\u_t, t\in[0,T]}\  & \int\limits_0^T \ell(\q_t,\u_t) {\rm d}t + \ell_f(\q_T)  & \\ \text{s.t. } \quad & \H(\q_t)\ddot{\q}_t + \C(\q_t,\dot{\q}_t)  =\B \u_t, &\forall t\in [0,T],\label{eq:oc_planning_stance_dc}\\ &  \q_t \in \Q, \ \u_t\in \U, & \forall t\in [0,T],\label{eq:oc_planning_stance_sc} }}
where $\ell_f(\q_T)$ is the terminal cost that incorporates the desired target state, $\ell(\q_t,\u_t)$ constitutes the running cost penalizing the state and input trajectories achieving the target state, $\U$ represents the input constraint set, and $\Q$ denotes the set of potential state constraints imposed on the system states.

\begin{remark}\label{rem:takeofflanding}
The above optimal control problem unifies the planning problems during both taking off and landing stages. The terminal state $\q_T$ in the terminal cost $\ell_f(\q_T)$ serves as the key term distinguishing these two problems. In specific, the terminal state is chosen to be a valid take-off state for planning during taking off, and is chosen to be a stably moving forward state for planning during landing.
\end{remark}

Due to the nonlinear and underactuated nature of the W-SLIP dynamics~\eqref{eq:oc_planning_stance_dc}, directly solving the above optimal control problem is challenging. To address these issues, we first show that the W-SLIP model inherits a nice property which helps in designing tractable planning algorithms.

\subsubsection{Structural Property of W-SLIP Dynamics}
Differential flatness~\cite{Mellinger2011,Sreenath2013,Chen2020} is a powerful property that commonly leads to significant simplification in control of nonlinear systems. However, based on the detailed form of~\eqref{eq:eom_wslip}, it is conjectured that the full dynamics for the W-SLIP model~\eqref{eq:eom_wslip} are not differentially flat, which prevents us from directly applying flatness based planning strategies. Nonetheless, we show that the dynamics of the W-SLIP model can be further simplified and differential flatness continues to contribute in constructing tractable algorithms. 
The first key feature we note regarding the W-SLIP model is that all $\H(\q)$, $\C(\q,\dot{\q})$ and $\B$ matrices in ~\eqref{eq:eom_wslip} are all independent of $x$, implying weak coupling between dynamics of $(L,\theta)$ and that of $x$. In addition, once trajectories of $L$ and $\theta$ over time are specified, it follows from the first row of ~\eqref{eq:eom_wslip} that:
\eqn{\tau_w\!  =\! r  \left( {{m_b} \!+\! {m_w}} \right)\ddot x\! + \! r {m_b} (\sin \theta (\ddot L\!-\!L\dot{\theta}^2) \! +  \!\cos \theta (L\ddot{\theta}\!+\!2\dot{L}\dot{\theta})) .}
Plugging this into the last two rows of~\eqref{eq:eom_wslip} and taking $\hat{\q} = (L,\theta)$ as the new configuration variable and $\hat{\u} = (\ddot{x},\Delta L) = (u_x,u_L)$ as the new inputs, we have
\eq{\label{eq:eom_rd_model}  \begin{aligned}
& \left[ {\begin{array}{*{20}{c}}
{{m_b}}&0\\
{r{m_b}\sin \theta }&{{m_b}{L^2} + r{m_b}L\cos \theta }
\end{array}} \right]  \ddot{\hat{\q}} 
\\ &+ \left[ {\begin{array}{*{20}{c}}
{ - {m_b}L{{\dot \theta }^2} + {m_b}g\cos \theta  - {{\bar K}_s}\left( {{L^0} - L} \right)}\\
{2{m_b}\left( {L + r\cos \theta } \right)\dot L\dot \theta  -m_bL\sin\theta(g+r\dot{\theta}^2) }
\end{array}} \right] 
\\ &= \left[ {\begin{array}{*{20}{c}}
{ - {m_b}\sin \theta }&{{{\bar K}_s}}\\
{ - {m_b}L\cos \theta  - r\left( {{m_b} + {m_w}} \right)}&0
\end{array}} \right]\hat{\u}, \end{aligned} }
which is fully actuated and hence differentially flat. This flatness property allows for significant simplification in planning of the leg length $L$ and leg angle $\theta$. This observation allows us to reformulate the W-SLIP planning problem by focusing on the dynamics of $L$ and $\theta$, and to address the underactuation challenge via the dynamics of $x$ in an implicit manner. \begin{figure}[tbp!]
	\centering
	{\includegraphics[width=0.75\linewidth]{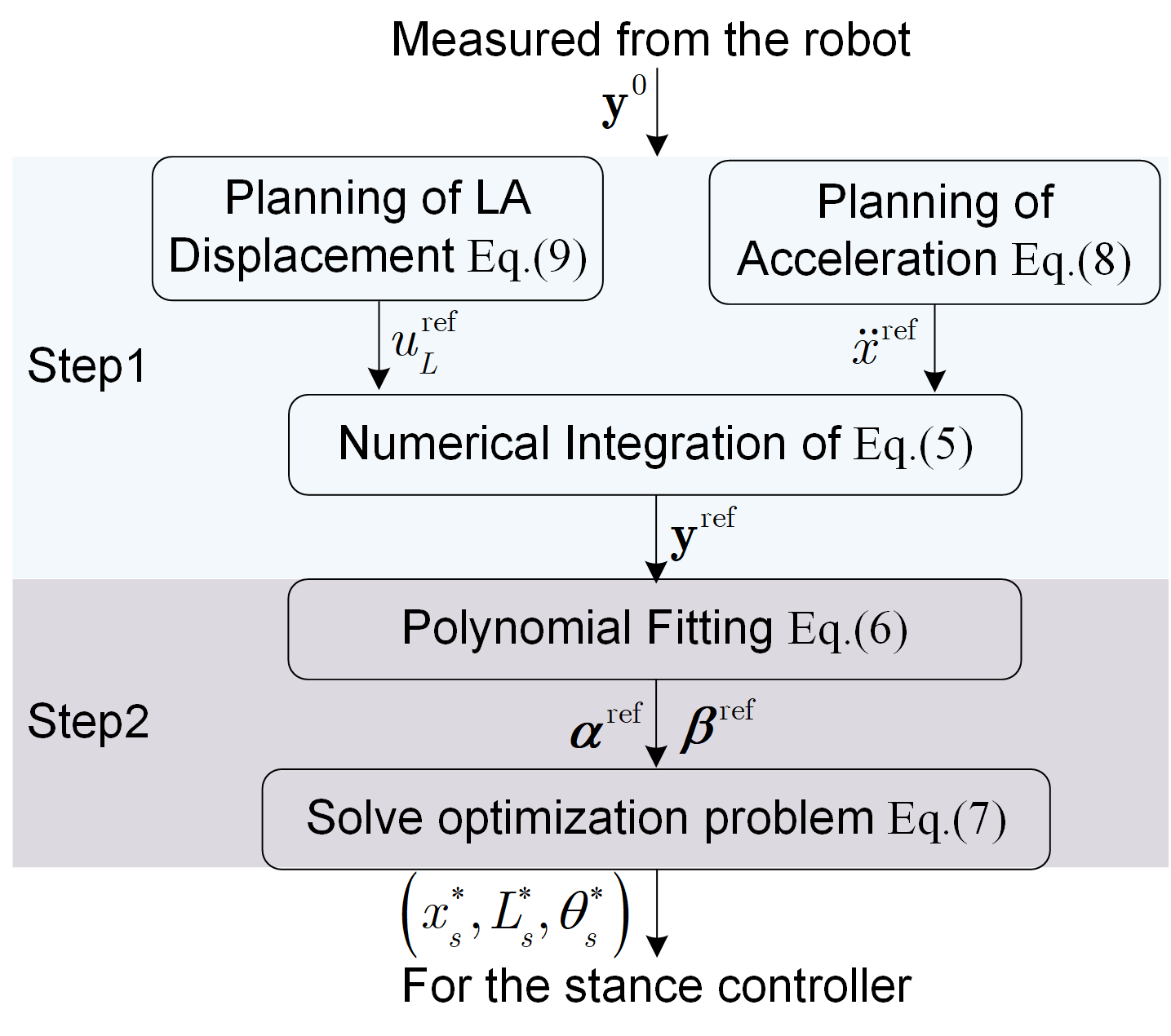}}
	\caption{\footnotesize Schematic diagram of flatness-based planning}
	\label{fig:stance_plan}
	\vspace{-15px}
\end{figure}

\subsubsection{Flatness-based stance phase planning}
Based on the above discussion, the planning problem for the W-SLIP can be simplified to be the planning of $L$ and $\theta$, which admits a quadratic programming based solution exploiting the differential flatness property. 

To begin with, we select the flat outputs of the dynamics~\eqref{eq:eom_rd_model} to be the configuration variables $L$ and $\theta$, and parametrize them with the following polynomial functions
\eq{\label{eq:polypara}  L(t) = \phi_0(t)\alpha,  \theta(t) = \phi_0(t)\beta,  \phi_0(t) = [1, t, t^2, \ldots, t^N],}
where $N$ is the degree of the polynomial parameterization, $\phi_0(t)$ is basis of the polynomial parameterization, $\alpha$ and $\beta$ are vectors of polynomial coefficients. Adopting this notation, it follows that the flat outputs and their derivatives can be collectively written as a linear function of the polynomial coefficients, i.e., $\y(t) = \bar{\Phi}(t) \gamma$ with $\gamma = (\alpha,\beta)$, $\y(t) = (L(t),\dot{L}(t),\ddot{L}(t),\theta(t),\dot{\theta}(t),\ddot{\theta}(t))$ and $\bar{\Phi}(t) = \blkdiag\{\Phi(t),\Phi(t)\}$ with \eq{\Phi(t) = \ar{{ccccc}1& t& t^2& \ldots& t^N \\ 0 & 1& 2t& \ldots & Nt^{N-1} \\ 0 & 0 & 2 & \ldots& N(N-1)t^{N-2}}.} As a result, planning of $L$ and $\theta$ can be formulated as the following quadratic programming over polynomial coefficients. 
\subeq{\label{eq:df_plan}\al{
		\min\limits_{\gamma  } \quad & \int\limits_0^{T} \| \bar{\Phi}(t)\gamma - \y^{\text{ref}}(t)\|_{Q_1}^2{\rm d}t  +  \|\bar{\Phi}(T)\gamma - \y^\text{d}\|_{Q_2}^2 \\
		\text{s.t. }\quad  & \bar{\Phi}(0) \gamma = \y^0.
}}	
where $\y^\text{d}$ is a vector containing the desired flat outputs at the end of the planning horizon obtained from $\q_T$ in the original problem, $Q_1$ and $Q_2$ are two weighting matrices tuning the weights of running and terminal costs, $\y^0$ is the flat output corresponding to the given initial state $(\hat{\q}_0,\dot{\hat{\q}}_0,\ddot{\hat{\q}}_0)$ in~\eqref{eq:oc_planning_stance}. 

\begin{remark}\label{rem:runningterminal}
$\y^\text{ref}(\cdot)$ in the above formulation plays a crucial role of implicitly addressing the state and input constraints that are involved in~\eqref{eq:oc_planning_stance}. By careful construction of $\y^\text{ref}(\cdot)$, these constraints can be incorporated in~\eqref{eq:df_plan} softly, promoting the resulting solution to respect these constraints.
\end{remark}

Construction of the reference trajectory $\y^\text{ref}(\cdot)$ involves two main steps, as depicted in Fig.~\ref{fig:stance_plan}. First, the input reference trajectories $\hat{\u}^\text{ref} = (\ddot{x}^\text{ref},u_L^\text{ref})$ are generated with a one-dimensional actuated spring-mass model and a simple double integrator model. Then, the reference trajectories $\y^\text{ref}(\cdot) = (L^\text{ref},\theta^\text{ref})$ are generated by forward simulating the nonlinear dynamics~\eqref{eq:eom_rd_model} with  $\hat{\u}^\text{ref}$. 

The forward acceleration reference $\ddot{x}^\text{ref}$ is obtained by solving the following optimal control problem with a simple double integrator model:
\begin{equation}\label{eq:dxref}
\begin{array}{l}
\mathop {\min   }\limits_{\ddot x^\text{ref}(\cdot)}  \displaystyle\int\limits_0^T {\ddot x^\text{ref}{{\left( t \right)}^2}{\rm d}t} \\
{\rm s.t.}\quad\dot x^\text{ref}\left( 0 \right) = {v_0},\dot x^\text{ref}\left( T \right) = {v_{T}},\ddot x^\text{ref}\left( T \right) = {{\dot v}_{T}}
\end{array}
\end{equation}
where $v_0$ is initial velocity, $v_{T}$ is the desired velocity, and $\dot{v}_{T}$ is the desired acceleration. By parameterizing $x(t)$ using a polynomial function, the above problem can be efficiently solved via standard quadratic programming techniques.

Following this idea, the linear actuation reference profile $u_L^\text{ref}$ is generated via solving an optimal control problem with the following one-dimensional actuated spring-mass model
\eqn{{m_b}\ddot L =  - {m_b}g + {\bar K_s}\left( {{L^0} - L + {u_L}} \right),}
and the corresponding optimal control problem is given by
\begin{equation}\label{eq:dlref}
\begin{array}{l}
\mathop {\min  }\limits_{{{u}^\text{ref}_L}(\cdot)} \displaystyle\int\limits_0^T {\ddot L^\text{ref}{{\left( t \right)}^2}{\rm d}t} \\
\begin{aligned}
{\rm s.t.} \quad {m_b}\ddot L^\text{ref} &=  - {m_b}g + {{\bar K}_s}\left( {{L^0} - L^\text{ref} + {u_L^\text{ref}}} \right)\\
L^\text{ref}\left( 0 \right) &= {L_0},\dot L^\text{ref}\left( 0 \right) = {{\dot L}_0}\\
L^\text{ref}\left( T \right) &= {L_{T}},\dot L^\text{ref}\left( T \right) = {{\dot L}_{T}},\ddot L^\text{ref}\left( T \right) = {{\ddot L}_{T}}\\
{u_{L\min }} &\le {u_L^\text{ref}} \le {u_{L\max }}
\end{aligned}
\end{array}
\end{equation}
where $(L_{T},\dot{L}_{T},\ddot{L}_{T})$ specifies the target state and ${u_{L\min }},{u_{L\max }}$ accounts for the input constraint in the original problem. Due to linearity of the actuated spring-mass model, a polynomial parameterization of $L(t)$ is also applicable, yielding a quadratic programming problem.

As mentioned before, once the above two problems~\eqref{eq:dxref} and~\eqref{eq:dlref} are solved, the reference trajectory $\y^\text{ref}(\cdot) = (L^\text{ref},\theta^\text{ref})$ can be generated through numerically integrating the nonlinear system~\eqref{eq:eom_rd_model} with the input references $\hat{\u}^\text{ref} = (\ddot{x}^\text{ref},u_L^\text{ref})$. To take further advantage of the polynomial parameterization~\eqref{eq:polypara}, the resultant solution $\y^\text{ref}(\cdot)$ is fitted with a polynomial with degree $N$. This additional curve fitting procedure allows for reformulation of~\eqref{eq:df_plan} using a standard quadratic program, whose solution will be denoted by $(x^*_s,L^*_s,\theta^*_s)$ in the sequel.

The above proposed stance phase planning scheme only involves solving three quadratic programs (QPs), a numerical integration problem and a polynomial fitting problem. Due to the fact that optimization variables are coefficients of the polynomials and degrees of these polynomials are typically no larger than $15$, the aforementioned QPs in general have no more than $30$ variables, allowing for online implementation in practice. In the following section, the flight phase planning problem is considered, which accounts for the flight phase evolution of the W-SLIP dynamics in a jump task. 

\section{Planning for Flight Phase}\label{sec:flightplanning}

The W-SLIP model during flight phase has a total of $4$ degrees of freedom, namely the two dimensional center of mass (CoM) coordinates in the sagittal plane, the leg angle $\theta$, and leg length $L$. During flight, gravity is the only external force applied to the W-SLIP model. As a result, the CoM position follows a ballistic trajectory and angular momentum of the entire system is conserved, i.e.,
\eq{\ \frac{d}{dt} (\I \dot{\theta}) = 0 , \text{ with }\I = \frac{m_b m_w}{m_b+m_w}{L^2} .}

Due to these restrictions, we have only one remaining degree of freedom kinematically for flight phase planning. Without loss of generality, let $(c_{x,0},\dot{c}_{x,0},c_{z,0},\dot{c}_{z,0},\theta_0,\dot{\theta}_0,$ $L_0,\dot{L}_0)$ be the initial state at take-off. Then the time corresponding to the apex CoM state is $T_{top} =  \dot{c}_{z,0} \slash g$.

Based on the previous discussions that the leg length $L$ is the only degree of freedom available for planning during flight phase, the planning problem for flight phase is simply formulated as follows

\subeq{\label{eq:flightplan}\al{\min\limits_{L(\cdot)}\quad   & \|L(T_{top})-L_\text{des}\|_2^2  \\  \text{s.t. } \quad & L(0) = L_0, \ \dot{L}(0) = \dot{L}_0, \\ &  L(T_f)  = L_{TD}, \dot{L}(T_f) = \dot{L}_{TD}, }} where $L_\text{des}$ is the desired leg length at the apex state.

The above optimization directly works with the leg length, whose objective is to minimize the leg length at the apex point thus creating large enough clearance. Constraints involved in the optimization problem are initial and terminal restrictions imposed on the leg length. In the above formulation, the overall time horizon $T_f$ is determined by the following relationship 
\eqn{\frac{m_b}{m_b+m_w} L_{TD} \cos(\theta_{TD}) = c_{z,0} + \dot{c}_{z,0}T_f-\frac{1}{2}gT_f^2,} whose solution is simply given by
\eqn{{T_f} \!= \! \frac{1}{g}\left( {{{\dot c}_{z,0}} \!+\! \sqrt {\dot c_{z,0}^2\! -\! 2g\left( \frac{m_bL_{TD}\cos( \theta_{TD})}{m_b+m_w} \!-\! c_ {z,0} \right)} } \right).}

By parametrizing the leg length profile with polynomials given by~\eqref{eq:polypara}, the flight phase planning problem is recast as quadratic programming over polynomial coefficients similar to as in the previous section. 
\section{Tracking Controllers}\label{sec:trackingcontroller}
The previous two sections mainly focus on the planning of feasible jumping motion with wheeled-bipedal robots. Once the planned trajectories are constructed, suitable controllers need to be applied for generating appropriate joint torques for the full order system. Throughout this section, we focus on the original three-dimensional model~\eqref{eq:wholebody} that requires consideration of the two wheels separately. 
  
\begin{figure}[t!] 
\centering
\subfigure[\footnotesize T-WIP model ]{%
\includegraphics[width=0.3\linewidth]{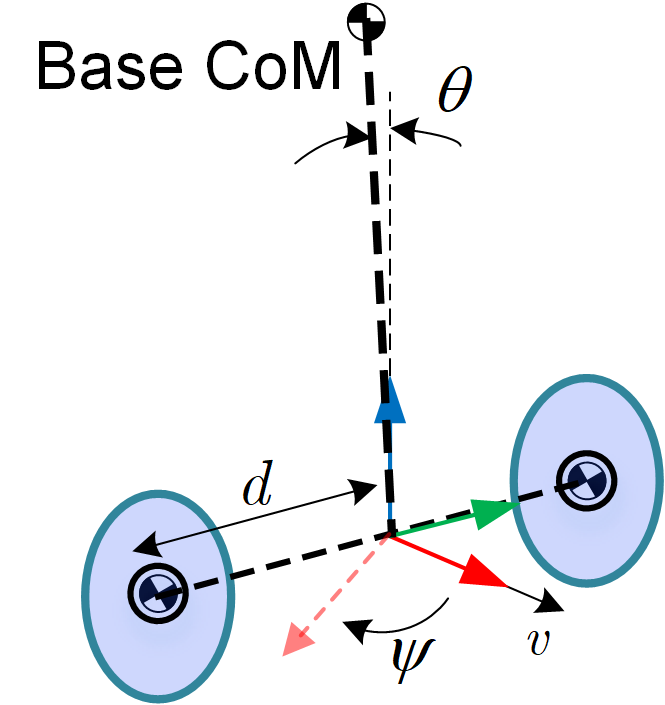}
\label{fig:twip}}\qquad \qquad 
\subfigure[\footnotesize Upper-body model]{%
\includegraphics[width=0.3\linewidth]{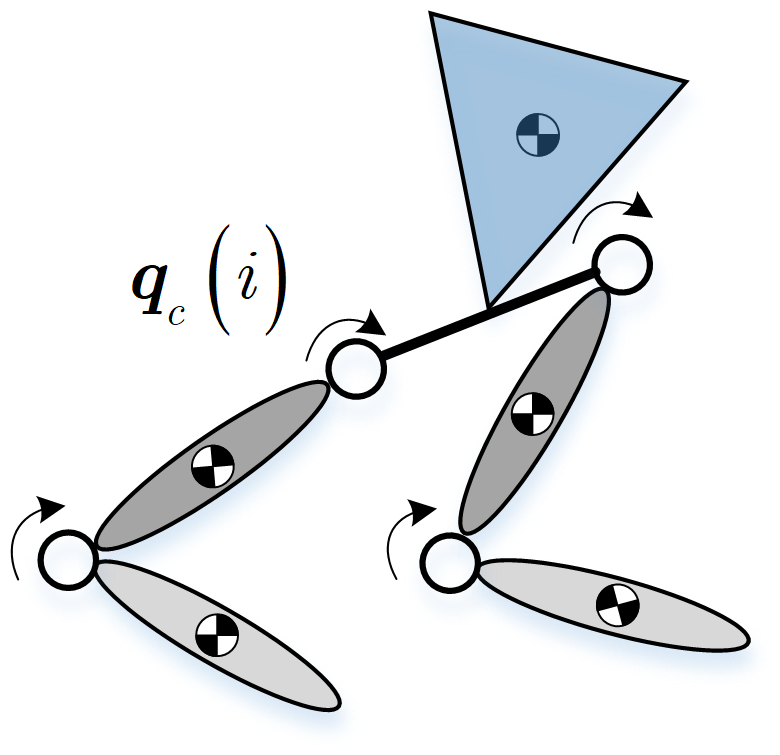}
\label{fig:mrigid}}
\caption{\footnotesize The two-wheeled inverted pendulum (T-WIP) model and the floating-base upper body model for controller design. }
\label{fig:models4controller}
\vspace{-15px}
\end{figure}

To begin with, we first develop a tracking controller for stance phase. To address the underactuated dynamics that prevent direct application of classical whole-body controllers~\cite{Kuindersma2016,Wensing2017}, we propose a composite controller that separately determines the joint torques at the wheels and at the upper body joints. In particular, the whole-body dynamics~\eqref{eq:wholebody} are approximated by a two-wheeled inverted pendulum (T-WIP) model (Fig.~\ref{fig:twip}) and a floating base $5$-link multi-rigid body system with $4$ actuators (Fig.~\ref{fig:mrigid}). Coupling terms between these decoupled models are estimated via disturbance observers, which enhances the tracking performance. 

The T-WIP model has been studied recently in the literature~\cite{Zhou2016,Kim2017,Zhou2019b} and is used in our controller to take care of the torques at the wheels. 
The T-WIP dynamics adopted for the stance control problem involve four degrees of freedom, whose state is considered to be $\x_{c} = (\theta,\dot{\theta},v,\dot{\psi})$, where $\theta$ is the tilting angle of the T-WIP, $v$ is the forward velocity along the heading direction and $\dot{\psi}$ is the yaw rate of the base. The T-WIP is linearized by considering the error state $\e_{c} = \x_{c} - \x_{c}^\text{ref}$ from the reference trajectories $\x_{c}^\text{ref} = (\theta^*_s,\dot{\theta}_s^*,\dot{x}^*_s,0)$ obtained from the stance planning problem in Section~\ref{sec:stanceplanning}. The model for the tracking controller design for wheels is then given by 
\eq{\label{eq:twip}\dot{\e}_{c} = \A_c(\x_{c}^\text{ref},L^*_s,\ttau_w^*)\e +\B_c(\x_{c}^\text{ref},L^*_s,\ttau_w^*) (\Delta \ttau_w +\d_w) ,} where $\ttau_w^*$ denotes the optimal torques at the wheels from the planning solution, $\Delta \ttau_w$ gives additional torques to be injected into the wheels, and $\d_w$ denotes all unmodeled terms that are not captured by the linearized system. Tracking controller synthesis requires finding the appropriate $\Delta \ttau_w$ resulting in $\lim\limits_{t\to \infty} \e_{c}(t) = 0$. Note that this T-WIP based controller for the wheels accounts for tracking of both $\theta^*_s,$ and $\dot{x}^*_s$ components from the planned trajectory while tracking of $L^*_s$ is handled via a task-space whole-body controller. 
An approximated floating-base multi-link rigid-body model used for control of $L^*_s$ is simply given below
\eq{\label{eq:wbcmodel}\H_{c}(\q_c)\ddot{\q}_c +\G_c + \d_c = \boldsymbol{\tau}_c,} in which the matrix $\H_c(\q_c)$ is approximated as a diagonal inertia matrix, $\G_c$ is the gravitational term, and $\boldsymbol{\tau}_c$ is a vector of joint torques at the hips and knees. In this model, we assume that the dynamical coupling terms with the wheels, the Coriolis term affected by the leg tilting angle and all the other unmodeled dynamics in~\eqref{eq:wbcmodel} are jointly viewed as a disturbance term $\d_c$.

To synthesize the time-varying linear quadratic regulator (TV-LQR) controller for the wheels and the whole-body controller for the hips and knees, the coupling terms $\d_w$ and $\d_c$ in~\eqref{eq:twip} and~\eqref{eq:wbcmodel} first need to be estimated. In this paper, we adopt two disturbance observers to estimate the disturbances $\d_w$ and $\d_c$ which will be compensated in the resulting controller. Inspired by the Super-Twisting-Algorithm~\cite{Moreno2012}, dynamics of the disturbance observers are designed as follows 
\subeq{\label{eq:dob}\al{ \dot{\xxi}_1 & = \A_c  \e_c +\B_c\Delta \ttau_w +\D_w  \\ \dot{\xxi}_2 & = \H_c^{-1}({ \ttau}_c - \G_c)+\D_c}} where $\xxi_1$ and $\xxi_2$ are observer states representing estimates of $\e_c$ and $\dot{\q}_c$, and $\D_w = \B_c\hat{\d}_w$ and $\D_c =  -\H_c^{-1}\hat{\d}_c$ are the quantities to be estimated that are constructed as follows \eqn{\ald{  \D_w &=  -K_1\sqrt{\lvert\xxi_1 - \e_c\rvert} \sigma(\xxi_1 - \e_c) - K_2 \int \sigma(\xxi_1 - \e_c) \dt  \\ \D_c & =  -K_3\sqrt{\lvert\xxi_2 - \dot{\q}_c\rvert} \sigma(\xxi_2 -\dot{\q}_c) - K_4 \int \sigma(\xxi_2 - \dot{\q}_c) \dt }} where $\sigma(\cdot)$ is the sign function.

With the estimated $\D_w$ and $\D_c$ from the above observers, the torque profile at the wheel actuators is determined by \eq{\label{eq:wheelcontrol} \ttau_w = \ttau_w^* + \Delta \ttau_w,} where $\Delta \ttau_w = -\boldsymbol{K}_\text{TVLQR} \e_c - \B_c^\dagger\D_w $ with $-\boldsymbol{K}_\text{TVLQR} \e_c $ being the feedback term resulted from the TV-LQR design and $- \B_c^\dagger\D_w$ being the term compensating the coupling terms. 

The torques injected into the hip and knee joints are determined via solving the following task-space control problem
\subeq{\label{eq:stancetrackwb}\al{ \min\limits_{{\bm \tau}_c , \ddot{\q}_c} & \frac{1}{2}\|\J_t\ddot{\q}_c + \dot{\J}_t\dot{\q}_c - \dot{\boldsymbol{r}}_{t,c} \|_{\bm Q}^2  \\ \text{ s.t.} \quad &  \H_c \ddot{\q}_c + \G_c -\H_c\D_c = \ttau_c \\ & \ttau_c \in [\underline{\ttau},\overline{\ttau}] }} where $\J_t$ is the task-space Jacobian, $\dot{\boldsymbol{r}}_{t,c} = (\0, \ddot{\p}_c)$ is the command for the task-space motion obtained through the following PD law with the stance phase planning result.
\eq{\label{eq:stancecontrolPD}\ddot{\p}_c = \ddot{\p}^*_s +\boldsymbol{K}_{P,s}(\p^*_s -  \p_s)+\boldsymbol{K}_{D,s}(\dot{\p}^*_s  - \dot{\p}_s).}

During flight phase, the tracking controller tracks the optimal leg length profile $L_f^*$ obtained from the flight phase planning problem~\eqref{eq:flightplan} discussed in Section~\ref{sec:flightplanning}. Such a controller has a similar form to that of the task-space whole-body controller~\eqref{eq:stancetrackwb} in the stance phase tracking controller, simply by replacing $\p_s^*$ in~\eqref{eq:stancecontrolPD} with $\p_f^*$ corresponding to $L_f^*$.

With the tracking controllers proposed in this section, the whole-body wheeled-bipedal robot is capable of tracking the planned trajectory in order to achieve a successful jump as demonstrated in the following section.

\section{Simulation Validations}\label{sec:simu}

\subsection{Simulation Platform}
The overall planning and control framework is validated using the open-source V-REP simulator based on a prototype wheeled-bipedal robot as shown in Fig.~\ref{fig:nezha}. The total mass of the robot is $12.7 \text{kg}$. Joint torque limits for hips and knees are set to be $60 \text{Nm}$ and for wheels are set to be $10 \text{Nm}$, which accurately represent the actuator torque limits of the motors used in the real robot. In addition, it is worth mentioning that the actuators used for the real hardware (Fig.~\ref{fig:nezha}) follows the proprioceptive design paradigm~\cite{Wensing2017b} with large diameter motors ($\SI{98}{mm}$) and low gear reduction ratios ($6:1$). Such an actuator design paradigm ensures back drivability that protects the actuators at the time of landing, and high torque-delivery bandwidth that enables realizing desired torque profiles.


\subsection{Performance of the Stance Planning Scheme}
Model parameters for the W-SLIP model used in stance planning are selected as follows. The mass for the floating base is set to be $m_b = 9.5 \text{kg}$, and the mass for the wheel is assumed to be $m_w = 3.2 \text{kg}$, which approximately represents the distribution of the mass of the real robot. The rest length for the spring is set to be $L^0 =  0.79\text{m}$ compatible with the real robot, and spring stiffness is chosen to be $\bar{K}_s = 418 \text{N}\slash\text{m}$, which is estimated via the maximum output torque of the real actuators and the leg length geometry of the real SUSTech Nezha (Fig.~\ref{fig:nezha}) following the idea discussed in~\cite{Xiong2018}. The planning scheme is re-solved every $40 \text{ ms}$ (i.e., at $25 \text{ Hz}$).
\begin{figure}[tp!]
\centering
{\includegraphics[width=0.85\linewidth]{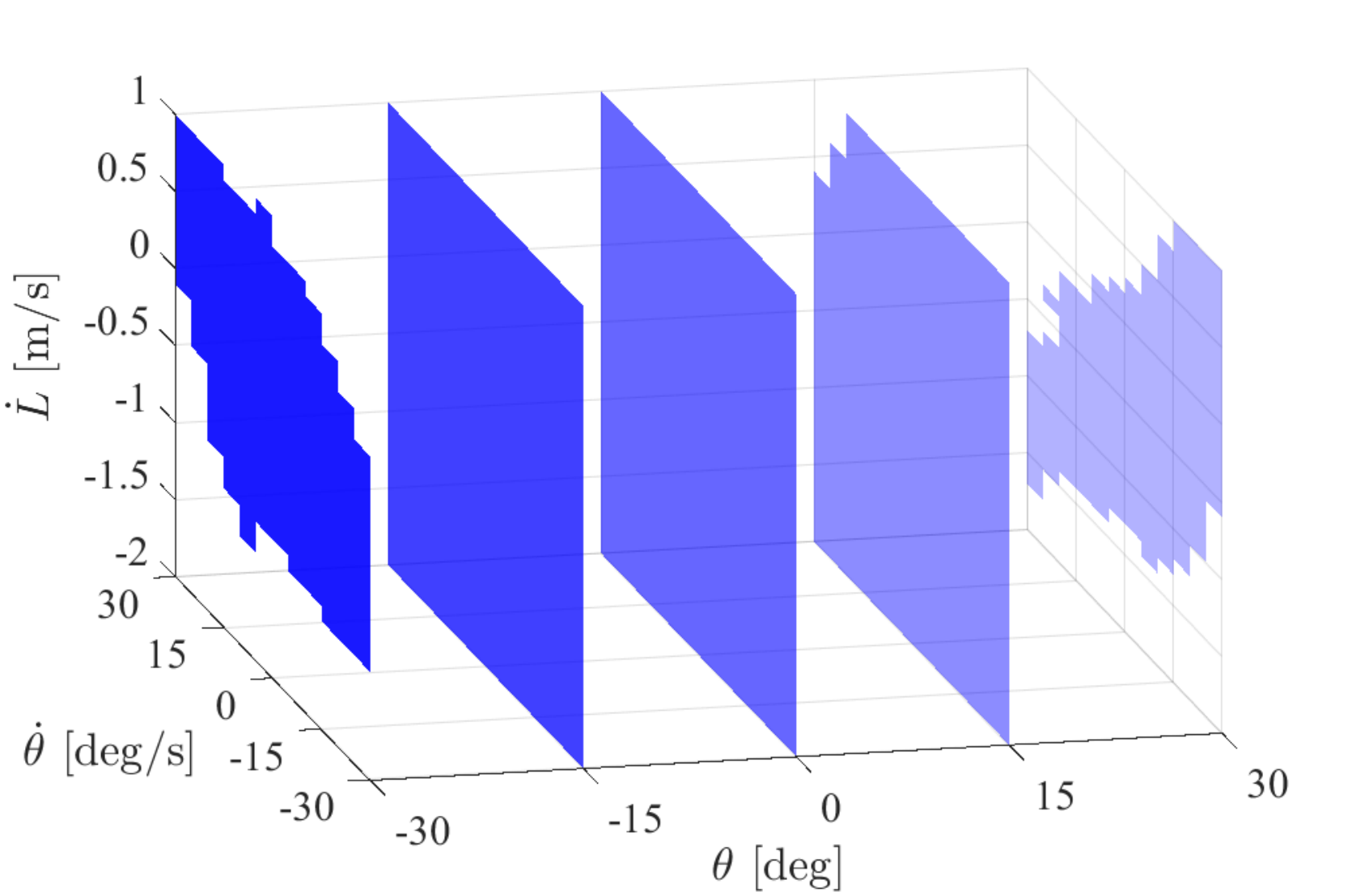}}
	\caption{\footnotesize  Region-of-attraction (RoA) over $(\dot{\theta},\dot{L})$. Slices with $\theta = \ang{-30}$, $\ang{-15}$, $\ang{0}$, $\ang{15}$, and $\ang{30}$ are shown. To approximately consider torque limits associated with SUSTech Nezha (Fig.~\ref{fig:nezha}), the maximum allowable torque at the wheel and maximum allowable force along the leg of the simplified W-SLIP model are set to be $\SI{12}{Nm}$ and $\SI{200}{N}$, respectively. }
\label{fig:RoAs}
	\vspace{-15px}
\end{figure}
Effectiveness of the proposed stance phase planning scheme is demonstrated with a desired take-off state $(\theta_{TO},\dot{\theta}_{TO},L_{TO},\dot{L}_{TO})\! =\! ( \ang{5},\ang{-9.91}\SI{}{\per s},\SI{0.79}{m}, \SI{3.42}{m \per s})$ which corresponds to a CoM apex with height $\SI{1}{m}$. Initial conditions for the horizontal position and horizontal velocity are specified to be $\SI{0}{m}$ and $\SI{0}{m \per s}$. Since there are still four dimensions remaining, we further restrict the leg length $L_0 = \SI{0.6}{m})$ and analyze the region-of-attraction (RoA) over the other three dimensions $(\theta,\dot{\theta},\dot{L})$, with $\theta$,  $\dot{\theta} $ and $\dot{L}$ ranging in$[\ang{-30},\ang{30}]$, $[\ang{-30}\SI{}{ \per s},\ang{30}\SI{}{ \per s}]$ and $[\SI{-2}{m \per s},\SI{1}{m \per s}]$. Fig.~\ref{fig:RoAs} shows five slices of the RoA over $(\dot{\theta},\dot{L})$ with $\theta = \ang{-30},\ang{-15},\ang{0},\ang{15},\ang{30}$. From this figure, it can be seen that the proposed planning scheme covers a wide range of the selected portion of state space. More specifically, the proposed planner fails when the tilting angle, the tilting rate and the leg length rate are all relatively large. Such a observation is consistent with intuition that if the W-SLIP is tilted too much with a large tilting rate as well, it is in general hard to accomplish the desired jumping with the limited control authority.  

\subsection{Performance of the Overall Scheme}
The overall planning and control scheme is validated through numerous V-REP simulations with SUSTech Nezha on a laptop with an Intel i5-9300H processor @ $\SI{4.10}{GHz}$ and $\SI{8}{Gb}$ memory. The robot is initialized at a vertically balanced configuration with floating base height at $\SI{0.79}{m}$.

\begin{figure}[t!] 
\centering
\subfigure[\footnotesize Time-lapse snapshots of V-REP simulation]{%
\includegraphics[width=0.9\linewidth]{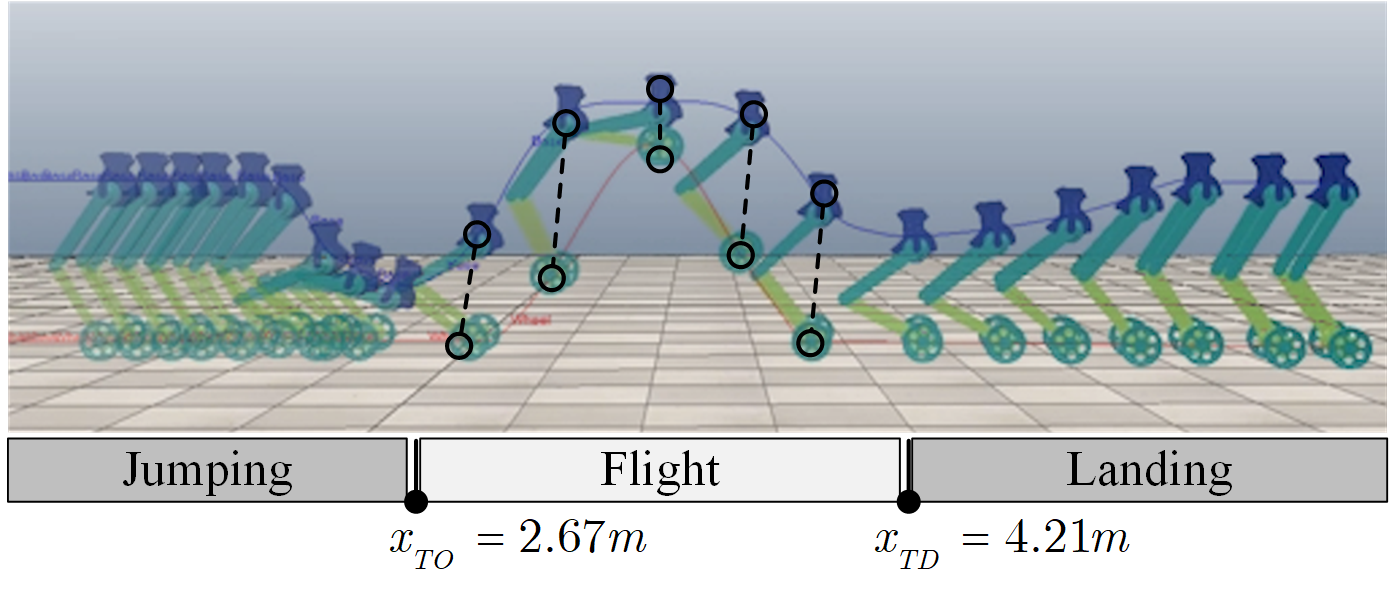}\label{fig:timelapse}}
\subfigure[\footnotesize Trajectories of floating base and wheel center. Part shaded in light blue corresponds to the above snapshots.]{%
\includegraphics[width=0.9\linewidth]{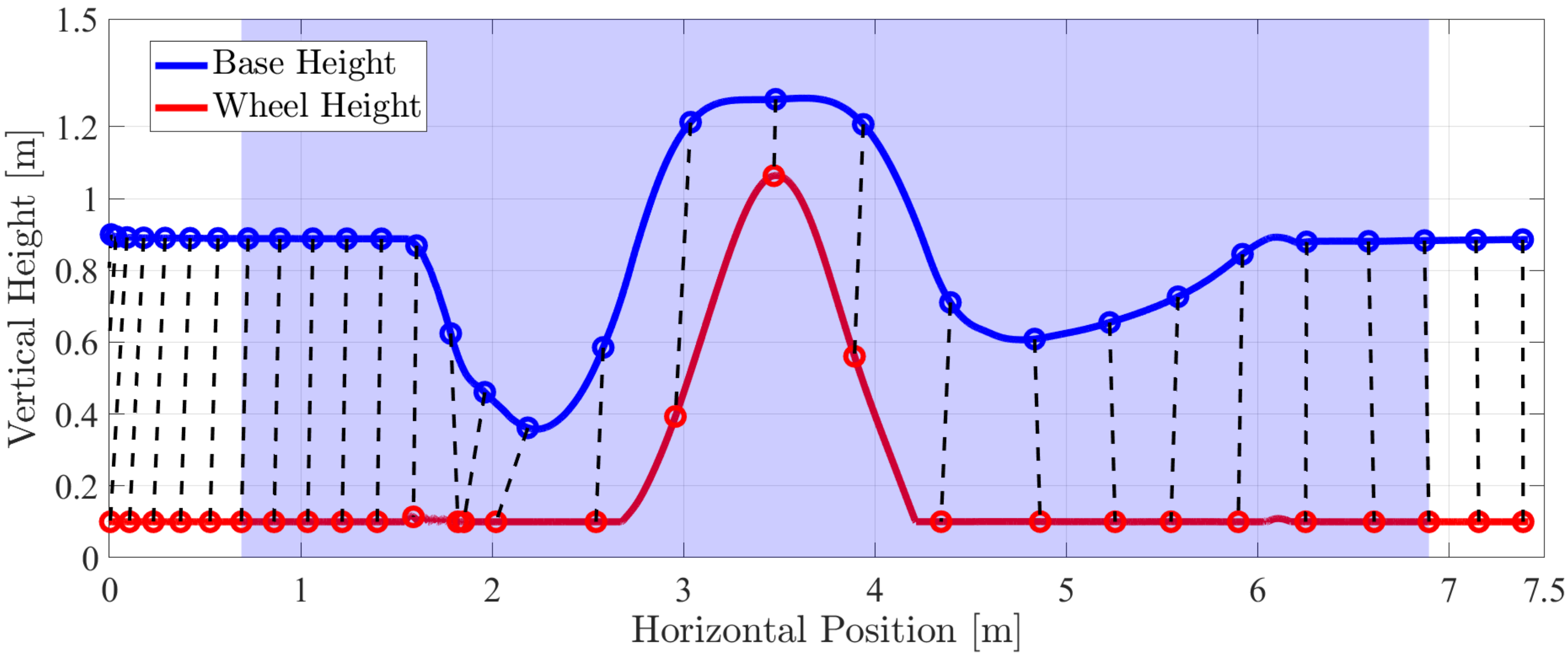}\label{fig:10_14_traj}}
\caption{\footnotesize   Jumping with SUSTech Nezha using the overall planning and control framework in V-REP simulator. Sampling frequency is $\SI{10}{\Hz}$ in both figures.}
\label{fig:jump14}
	\vspace{-15px}
\end{figure}
\begin{figure}[bp!]
	\centering
	{\includegraphics[width=\linewidth]{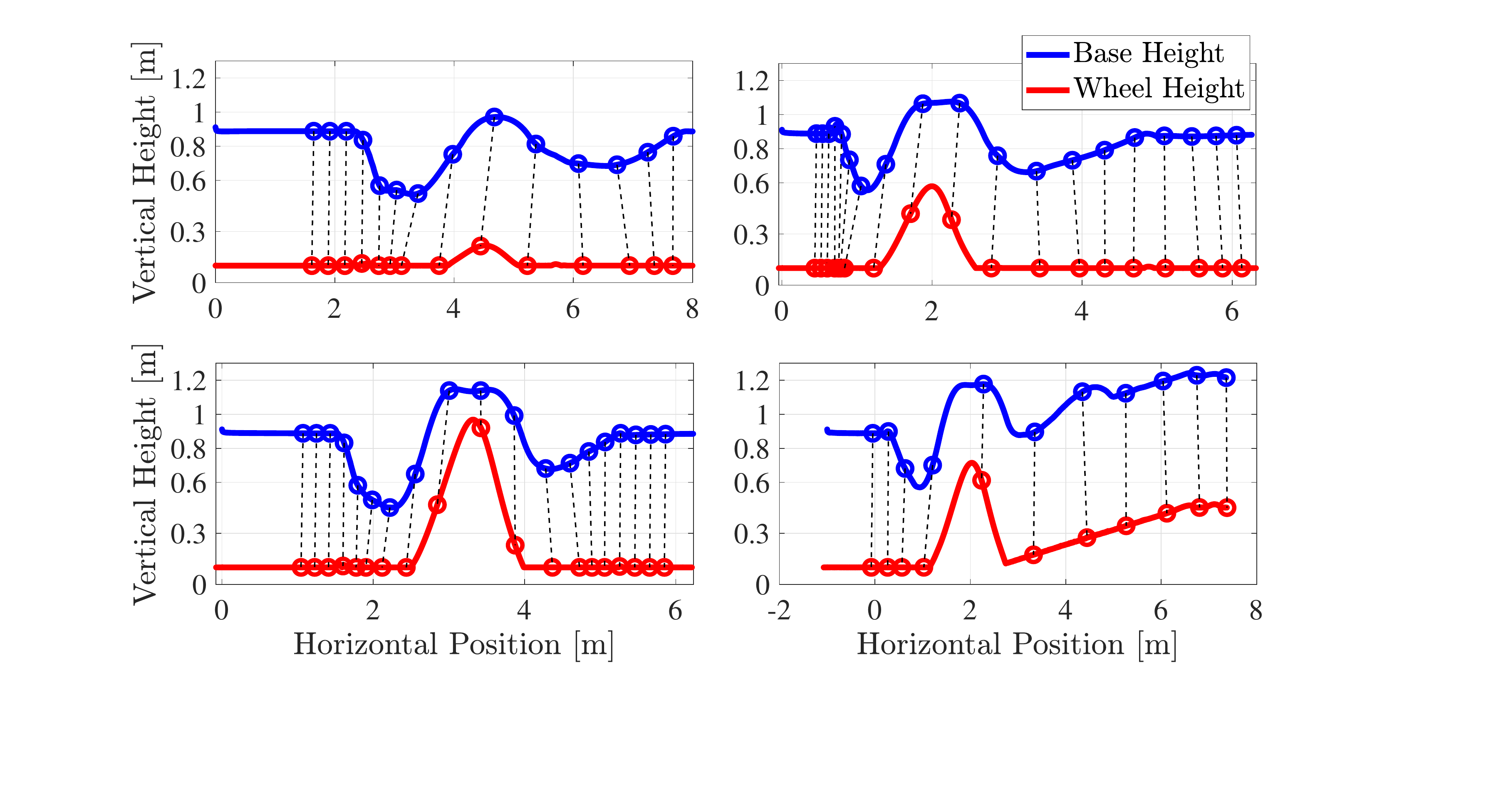}}
	\caption{\footnotesize  Four different successful jumps with different initial velocity and desired CoM apex height using the proposed framework. Top-left: CoM apex: $\SI{0.8}{m}$, initial forward velocity: $\SI{1.5}{m\per s}$. Top-right: CoM apex: $\SI{1}{m}$, initial forward velocity: $\SI{0.5}{m\per s}$. Bottom-left: CoM apex: $\SI{1.2}{m}$, initial forward velocity: $\SI{1}{m\per s}$. Bottom-right: landing on a slope with inclination $\ang{5}$.}
	\label{fig:jumps}
\end{figure}
In Fig.~\ref{fig:jump14}, time-lapse snapshots and evolution of the floating base and wheel positions are depicted for a successful jump with desired CoM apex height $\SI{1.4}{m}$. It should be noted that for this successful jump, the actual apex height achieved is lower than the desired one. This difference is attributed to the discrepancies between the W-SLIP and the multi-link robot. 

Fig.~\ref{fig:jumps} demonstrates more successful jumps achieved by the proposed strategy, with different commanded forward velocities ($\SI{0.5}{m \per s}$,$\SI{1}{m \per s}$ and $\SI{1.5}{m \per s}$) and different desired CoM apex heights ($\SI{0.8}{m}$, $\SI{1}{m}$ and $\SI{1.2}{m}$). Furthermore, a successful jump with landing on a slope with inclination $\ang{5}$ is also demonstrated, further validating the robustness of the proposed framework. In all simulation tests, switches to the stance planning for landing phase are triggered by the condition that velocity of the wheel is reset to zero ($\SI{0.05}{m \per s}$ is used as the tolerance in the implementation). The reader is kindly referred to the supplemental video for additional results.

To sum up, these simulations show that the proposed framework grants wheeled-bipedal robots the ability to jump with different heights at different velocities, enabling them to tackle challenging terrains such as wide gaps and high obstacles. 



\section{Concluding Remarks and Future Works}\label{sec:conclusion}

In this paper, a comprehensive hierarchical planning and control framework for jumping with wheeled-bipedal robots is developed. A novel wheeled-spring-loaded inverted pendulum (W-SLIP) model is proposed for the planning problems. Inspired by a differential-flatness-like property of the W-SLIP model, a quadratic programming based solution to the stance phase planning is devised. Planning during flight phase is addressed via another quadratic program with a kinematic model of W-SLIP. To accurately track the planned motion, a novel disturbance observer based composite controller for the stance phase and a standard task-space whole-body controller for the flight phase are devised. The performance of the overall framework is verified with V-REP simulation.

In the future, we aim to conduct experimental validations on hardware to further demonstrate the effectiveness of the proposed approach. Furthermore, extension of the proposed W-SLIP model to three dimensional cases is another important direction for future study.

\bibliographystyle{IEEEtran}
\bibliography{handle}

\end{document}